\documentclass[12pt]{article}
\usepackage{latexsym,amssymb}
\usepackage{multirow}
\usepackage{url}

\usepackage{amsmath, amsthm,epsfig,bm,booktabs,amssymb}
\usepackage{algorithm,algorithmic}
\usepackage{amsmath, amsfonts}
\usepackage[listformat=empty]{caption}
\usepackage{amsmath,latexsym,amssymb,epsf,amsfonts,amsthm}
\usepackage{geometry}
\usepackage{epsf}
\usepackage{color}
\usepackage{mathrsfs}
\usepackage{epstopdf}
\usepackage{threeparttable}
\usepackage{graphicx}
\usepackage{subfig}
\usepackage{tabularx}
\usepackage[all]{xy}

\newcommand{\w}{\mathbf{w}}

\def \R {{\mathbb R}}

\def \Q {{\mathbb Q}}

\newcommand{\lam}{\mbox{$\lambda$}}

\newcommand{\gam}{\mbox{$\gamma$}}
\newcommand{\nit}{\noindent}

\newcommand{\p}{\mathrm{proj}}
\newcommand{\be}{\begin{equation}}
\newcommand{\ee}{\end{equation}}
\newcommand{\ba}{\begin{eqnarray}}
\newcommand{\ea}{\end{eqnarray}}
\newcommand{\bi}{\begin{itemize}}
\newcommand{\ei}{\end{itemize}}

\newcommand{\comments}[1]{}

\begin{document}
\title{Median Binary-Connect Method and a Binary Convolutional Neural Nework for Word Recognition}

\vspace{2 in}

\author{Spencer Sheen \thanks{Department of Computer Science, UC San Diego, La Jolla, CA 92093.
 Email: spsheen97@gmail.com.}\;  and 
%\hspace{.1 in} \\
Jiancheng Lyu \thanks{Department of Mathematics, UC Irvine, Irvine, CA 92697. Email: jianchel@uci.edu.}}
%\date{December 14, 2017.}
\date{}
\maketitle
\thispagestyle{empty}
%\newpage 

\begin{abstract}
We propose and study a new projection formula for training binary weight convolutional neural networks. The projection formula measures the error in approximating a full precision (32 bit) vector by a 1-bit vector in the $\ell_1$ norm instead of the standard $\ell_2$ norm. The $\ell_1$ projector is in closed analytical form and involves a median computation instead of an arithmatic average in the $\ell_2$ projector. Experiments on 10 keywords classification show that the $\ell_1$ (median) BinaryConnect (BC) method outperforms the regular BC, regardless of cold or warm start. The binary network trained by median BC and a recent blending technique reaches test accuracy 92.4 \%, which is 1.1\%  lower than the full-precision network accuracy 93.5 \%. On Android phone app, the trained binary network doubles the speed of full-precision network in spoken keywords recognition. 

\end{abstract}
\bigskip

\hspace{.12 in} {\bf Keywords:}

\hspace{.12 in} Median Based Binarization

\hspace{.12 in} Closed Form Solution

\hspace{.12 in} Convolutional Neural Network

\hspace{.12 in} Spoken Keywords Classification.

\hspace{.12 in} Android App.

%\hspace{.12 in} {\bf AMS subject classifications:} 92B20, 65K10, 90C26.

\newpage 

\section{Introduction}
\setcounter{equation}{0}
Speeding up convolutional neural newtorks (CNN) on mobile platform such as cellular phones is 
an important problem in real world application of deep learning. The computational 
task is to reduce network complexity while maintaining its accuracy. In this 
paper, we study weight binarization of the simple audio CNN \cite{sp15} 
on TensorFlow \cite{tf18} for keyword classification, propose a new class of training algorithms, 
and evaluate the trained binarized CNN on an Android phone app. 
\medskip

The audio CNN \cite{tf18} has two convolutional layers and one fully 
connected layer \cite{sp15}. Weight binarization refers to restricting the weight values in each layer to 
the form $\pm \, {\rm sc}$ where $sc$ is a common scalar factor 
of full precision (32-bit). The network complexity is thus reduced with no change of 
architecture. The network training however 
requires a projection step and a modification of the classical 
projected gradient descent (PGD) called BinaryConnect \cite{bc_15}. If the projection is 
in the sense of Euclidean distance or $\ell_2$ norm, closed form solution is 
available \cite{xnornet_16}. With this $\ell_2$ projector, BinaryConnect (BC) updates 
both binary weight and an auxiliary float precision weight to avoid weight stagnation that occurs in 
direct PGD: $w^{k+1} = {\rm proj} (w^k - \eta \, \nabla f(w^k))$, $\eta > 0$ small, 
$f$ the objective function. In other words, the small variation of $w^k$ may vanish under the projection to 
discrete targets $\pm \, {\rm sc}$, rendering $w^{k+1}=w^{k}$. The BC update is: 
\be 
w^{k+1}_{f} = w^{k}_{f} - \eta\, \nabla f(w^k),\; w^k = {\rm proj} \, (w^{k}_{f}), \label{intro1}
\ee
where the float weight $w^k_{f}$ continues to evolve and eventually moves the binary weight $w^k$.      
Two improvements of BC stand out in recent works. One called BinaryRelax (BR) is to replace the ``hard 
projection'' to discrete targets $\pm \, {\rm sc}$ by a relaxed (approximate) projection 
still in analytical closed form \cite{BR_18} so to give $w^{k}$ and $w^{k}_{f}$ more freedom to 
explore the high dimensional non-convex landscape of the training loss function.   
The relaxation is tightening up as training goes forward and becomes a hard projection as 
the training reaches the end to produce a binarized output. BR outperforms BC on image 
datasets (CIFAR-10, Imagenet) and various CNNs 
%(VGG, ResNet) 
\cite{BR_18}. 
The other technique is ``blending'' \cite{BCGD_18}, which replaces $w^{k}_{f}$ in BC's gradient update by 
the convex linear combination $(1-\rho)\, w^{k}_{f} + \rho\, w^k$, for a small $\rho > 0$. 
With blending, the sufficient descent inequality holds at small enough $\eta$: 
$f(w^{k+1})-f(w^k) \leq - c\, \|w^{k+1}-w^{k}\|^{2}$, for some positive constant $c$, if $f$ 
has Lipschitz continuous gradient \cite{BCGD_18}.  
\medskip

In this paper, we replace the $\ell_2$ projector in BC \cite{bc_15} by a new projector in the $\ell_1$ sense and 
compare their performance on audio keywords data as well as CIFAR-10 image dataset. The $\ell_1$ projector is found 
in closed analytical form, and it differs from the $\ell_2$ projector in that the median 
operation replaces an arithmatic average operation. It is known that $\ell_1$ norm as a  
measure of error is more robust to outliers than $\ell_2$, see \cite{sp16} for a case study on 
matrix completion. Our experiments on TensoFlow show that BC with $\ell_1$ projector (called median BC) 
always outperforms BC under either cold (random) or warm start. 
Similar findings are observed on VGG network \cite{vgg_14} and CIFAR-10. 
\medskip

The rest of the paper is organized as follows. 
In section 2, we first present the $\ell_1$ projector, then summarize 
BC, median BC and BR algorithms as well as the audio CNN architecture.
In section 3, we compare test accuracies for 10 keywords classification and CIFAR-10 
on binary weight CNNs trained by the three algorithms. The best trained binary CNN and full precision audio CNN are implemented on an 
Android phone app.  The binary CNN doubles the processing speed of full-precision CNN on 
the app when words on the list are spoken and correctly recognized. 
The concluding remarks are in section 4.
\medskip

\section{Binary Quantization and Training Algorithms} 
\setcounter{equation}{0}

\subsection{Binary Weight Projections}
We consider the problem of finding the closest binary vector to a given 
real vector $w$, or the projection ${\rm proj}_{\Q} \, w$, for $w \in \R^D$, $\Q =\R_{+}\times \{\pm 1\}^D$. 
When the distance is Euclidean (in the sense of $\ell_2$ norm $\|\cdot \|$)), the problem:
\be
{\rm proj}_{\Q,a} (w):= {\rm argmin}_{z \in \Q}\; \|z - w\| \label{p1}
\ee
has exact solution \cite{xnornet_16}: 
\be
{\rm proj}_{\Q,a} (w) = {\| w \|_1 \over D} \; {\rm sgn}(w),  \label{p2}
\ee
where the $\ell_1$ norm $\| w\|_1:= \sum_{j=1}^{D} |w_j|$ and ${\rm sgn}(w)=(q_j)$, 
\[
 q_{j} = \left \{ \begin{array}{rr}
              1  & {\rm if}\;  w_j \geq 0 \\
             -1 &  \; {\rm otherwise}
          \end{array} \right.
\]
The solution is simply the sign of $w$ times the average of the absolute values of the components of $w$. 
\medskip

Now let us consider the distance in the $\ell_1$ sense, or solve:
\be
{\rm proj}_{\Q,m} (w):= {\rm argmin}_{z \in \Q}\; \|z - w\|_1.  \label{p3}
\ee
Write $z = s \, q$, where $s > 0$, $q=(q_j)$, $q_j= \pm 1$. Then:
\be
\| z - w \|_{1} = \sum_{j=1}^{D}\; | s \, q_j - w_j | \label{p4}
\ee
clearly $q_j = {\rm sgn}(w_j)$, and so:
\be
{\rm proj}_{\Q,m} (w)= s^* \, {\rm sgn}(w), \label{p5}
\ee
\be
 s^*={\rm argmin}_{s}\;  \sum_{j=1}^{D}\; \left |\; s - |w_j| \; \right | =
 {\rm median}(|w_1|, |w_2|, \cdots, |w_D|). \label{p6}
\ee
For a derivation of the median solution in (\ref{p6}), see section 2 of \cite{sp16} where a robust 
low rank matrix factorization in the $\ell_1$ sense is studied. 
 We shall call (\ref{p5})-(\ref{p6}) the median based binary projection. 

\subsection{Binary Weight Network Training Algorithms}
 Let the training loss function be $f$, and learning rate be $\eta$. We compare 3 algorithms below.
\medskip

\nit $\bullet$ Binary-Connect (BC \cite{bc_15}): 
\be
\w_f^{t+1} = \w_f^t - \eta \, \nabla f(\w^t), \; \w^{t+1} = \p_{\Q,a}(\w_f^{t+1}), \label{alg1}
\ee
where $\{\w^t\}$ denotes the sequence of desired quantized weights, 
and $\{\w^t_f\}$ is an auxiliary sequence of floating weights (32 bit). 
\medskip

\nit $\bullet$ Binary-Connect with median based quantization (Median BC):
\be
\w_f^{t+1} = \w_f^t - \eta \, \nabla f(\w^t), \; \w^{t+1} = \p_{\Q,m}(\w_f^{t+1}). \label{alg2}
\ee
\medskip

\nit $\bullet$ Binary-Relax (BR \cite{BR_18}): 
\be
\w_f^{t+1} = \w_f^t - \eta \, \nabla f(\w^t), \; \w^t = (\lam_t \, \p_{Q,a}(\w_{f}^{t})+ \w_f)/(\lam_t +1), \label{alg3}
\ee
where $\lam_t = \gam \lam_{t-1}$ for some $\gam > 1$.
\medskip

%\nit $\bullet$ Binary-Relax with median based %quantization (Median BR): 
%\be
%\w_f^{t+1} = \w_f^t - \eta \, \nabla f(\w^t), \; \w^t %= (\lam_t \, \p_{Q,m}(\w_{f}^{t})+ \w_f)/(\lam_t +1). %\label{alg4}
%\ee
%\medskip

In addition, we shall compare the blended version \cite{BCGD_18} of the 
above three algorithms obtained from replacing 
$\w_{f}^{t}$ in the gradient update by $(1-\rho)\, \w_{f}^{t} + \rho \, \w^t$, for $0<\rho \ll 1 $.  
\medskip

\subsection{Network Architecture}
Let us briefly describe the architecture of keyword CNN \cite{sp15,tf18} to 
classify a one second audio clip as either silence, an unknown word, `yes', `no', `up', `down', `left', `right', `on', `off', `stop', or `go'.
After pre-processing by windowed Fourier transform, the input becomes 
a single-channel image (a spectrogram) of size $t \times f$, 
same as a vector $v \in \R^{t\times f}$, where $t$ and $f$ are the input feature dimension 
in time and frequency respectively. Next is a convolution layer that operates as follows.
A weight matrix $W \in \R^{(m\times r)\times n}$ is 
convolved with the input $v$. The weight matrix is a  
local time-frequency patch of size $m\times r$, where $m \leq t$ and $r \leq f$. 
The weight matrix has $n$ hidden units (feature maps), and may down-sample 
(stride) by a factor $s$ in time and $v$ in frequency. The output of the 
convolution layer is $n$ feature maps of size 
$(t - m + 1)/s \times (f-r +1)/v$. Afterward, a max-pooling operation replaces 
each $p\times q$ feature patch in time-frequency domain by the maximum value, which 
helps remove feature variability due to speaking styles, distortions etc. 
After pooling, we have $n$ feature maps of size 
$(t - m + 1)/(sp) \times (f-r +1)/(vq)$. 
%An illutration is in Fig. \ref{conv}.
\medskip

%\begin{figure}
%%\begin{center}
%%\begin{tabular}{cc}
%\includegraphics[width=1.00\textwidth]{convlayer.png}
%%&
%%\includegraphics[width=0.48\textwidth]{No_1bit.png}\%\
%\end{tabular}
%\caption{A typical convolutional layer performing %convolution with stride, and max-pooling.} %\label{conv}
%\end{center}
%\end{figure}

The overall architecture \cite{tf18} consists of two convolutional layers, 
one fully-connected layer followed by a softmax function to output 
class probabilities at the end. The training loss is the standard cross entropy function. 
The learning rate begins at $\eta =0.001$, and is reduced by a factor of 10 in the 
late training phase. 
 \medskip

\section{Experimental Results}
In this section, we show TensorFlow training results of binary weight audio CNN 
based on 
%Binary-Connect
%/Binary-Relax 
%with median quantization (
median BC, BC and BR. 
With random (cold) start and 16000 iterations (default, about 70 epochs), 
BR is the best at 3.4 \% loss from the full precision accuracy. 
Increasing the number of iterations to 21000 from 16000 (warm start), we found that 
the full precision accuracy goes up to 93.5 \%, the highest level observed in our experiments.   
The binary weight training algorithms also improve, with BR and median BC in a near tie, 
at 1.4 \% (1.5\%) loss from the full precision accuracy respectively.   
For the same warm start, if blending is turned on ($\rho=10^{-5}$), 
median BC takes the lead at 92.4 \%, which narrows down the loss to 1.1 \% from the full precision accuracy, 
the best binary weight model that has been trained. Table \ref{ac1} shows all experimental 
results conducted in TensorFlow on a single GPU machine with NVIDIA GeForce GTX 745.
\medskip

We see from Table \ref{ac1} that BC benefits the most from blending while all methods 
improve with warm start. Median BC is better than BC in all cases. 
%On the other hand, median BR is not as good as BR %unless blending is introduced. 
%BR is the only method that gets worse under blending.
%\medskip
%
Since BR involves two stages of training and is more complicated to implement 
than BC's single stage training, median BC strikes a good balance of accuracy and  
simplicity for binary weight network training.    
 \medskip

\begin{figure}
\begin{center}
\begin{tabular}{cc}
\includegraphics[width=0.48\textwidth]{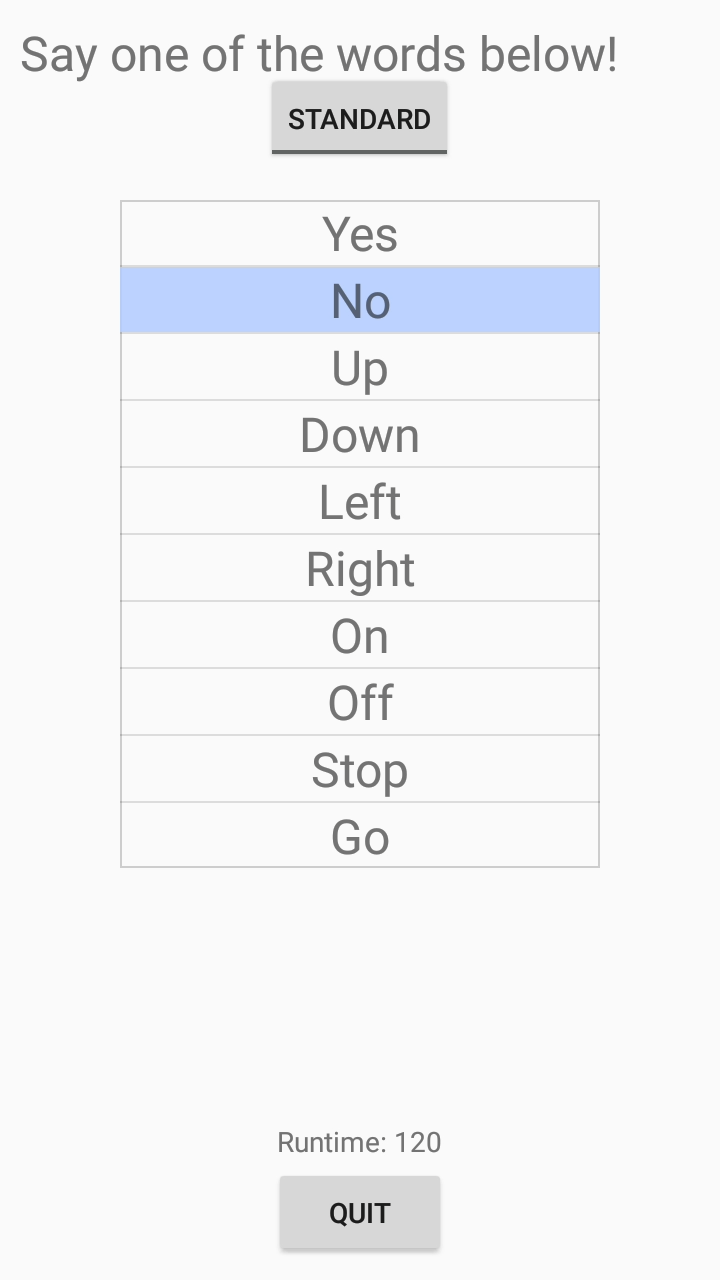}&
\includegraphics[width=0.48\textwidth]{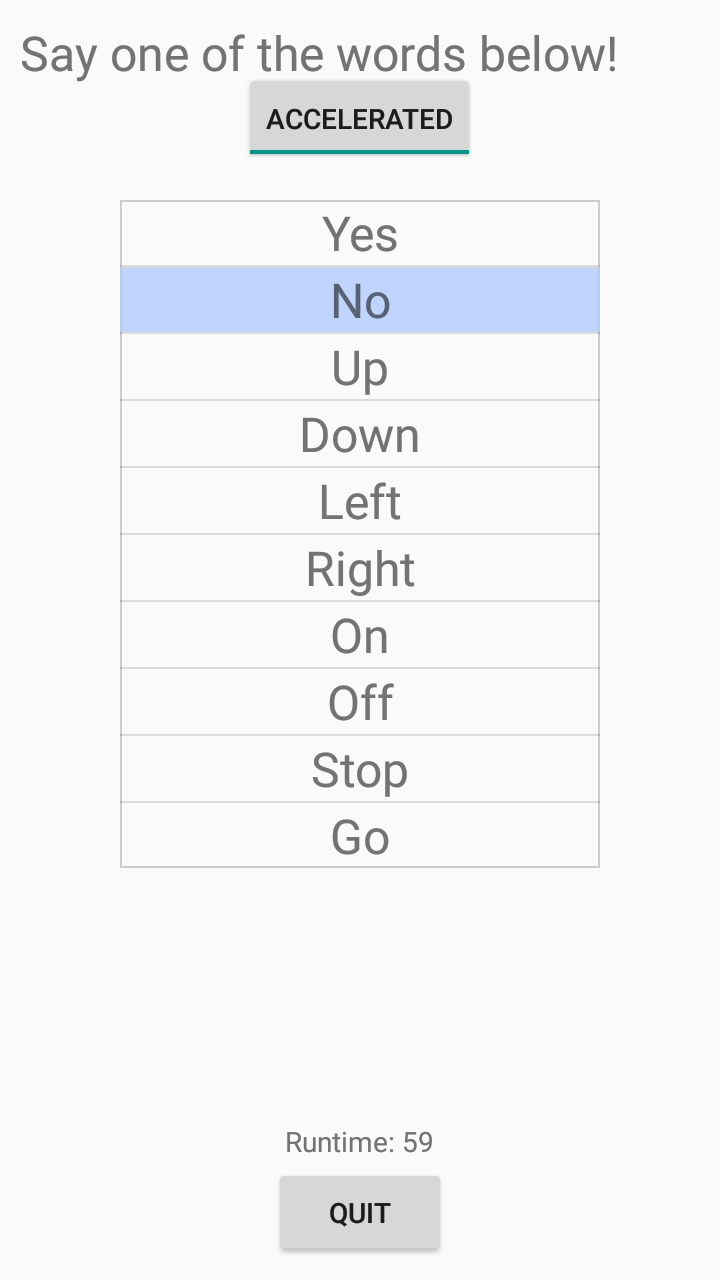}\\
\end{tabular}
\caption{Recognition of a spoken word `No' and runtimes (ms) on Android App. Left: full precision CNN model. 
Right: binary weight CNN model with 2x speedup.} \label{nodemo}
\end{center}
\end{figure}

\begin{figure}
\begin{center}
\begin{tabular}{cc}
\includegraphics[width=0.48\textwidth]{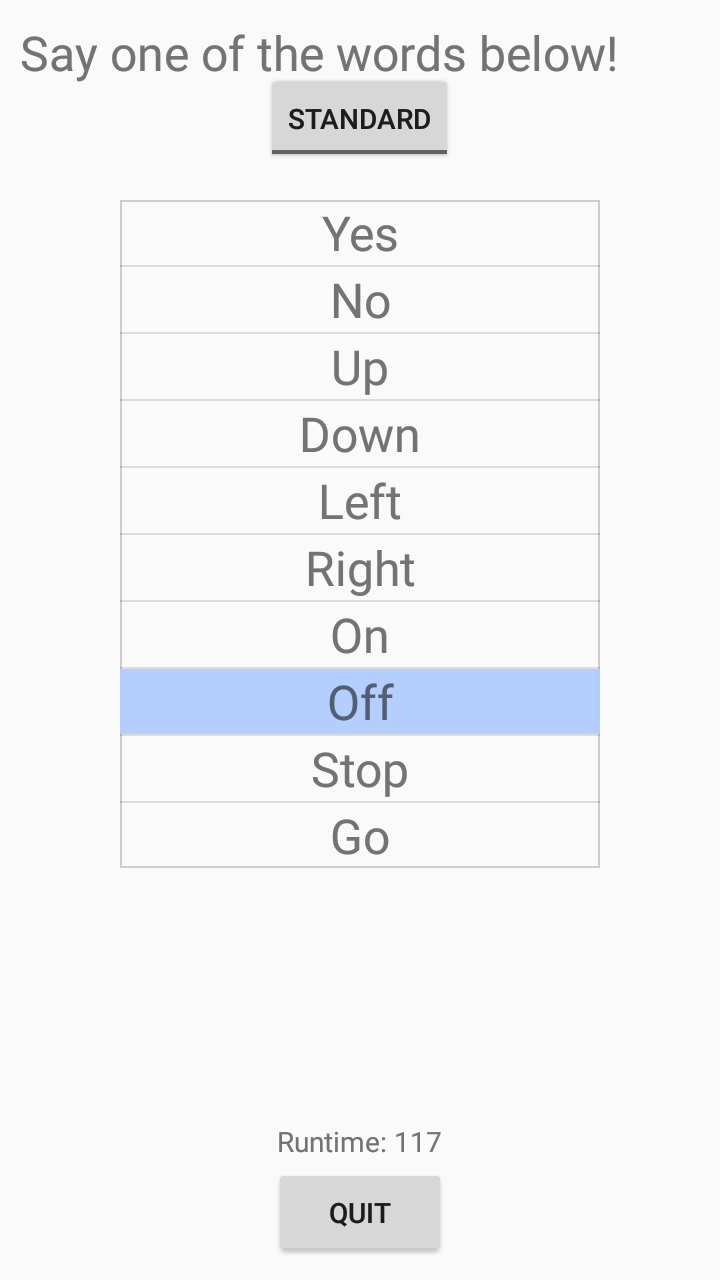}&
\includegraphics[width=0.48\textwidth]{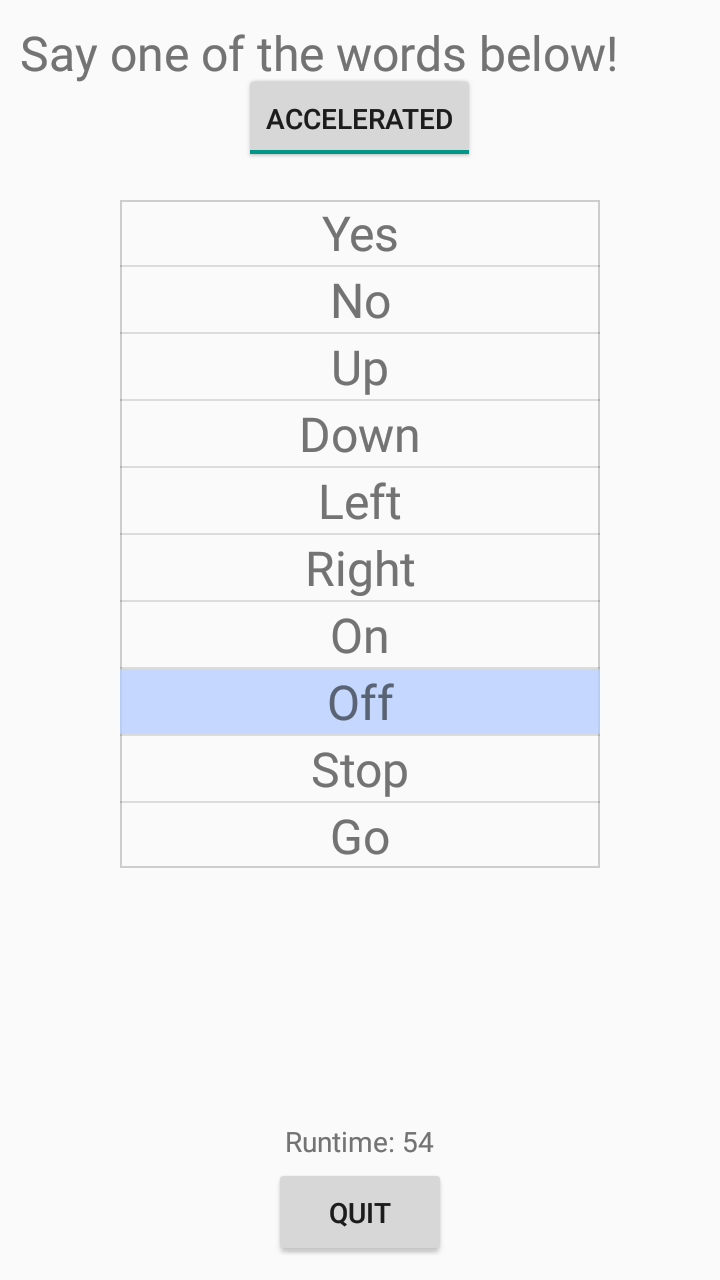}\\
\end{tabular}
\caption{Recognition of a spoken word `Off' and runtimes (ms) on Android App. Left: full precision CNN model. 
Right: binary weight CNN model with 2x speedup.} \label{offdemo}
\end{center}
\end{figure}

The most accurate full precision (32-bit) and binary (1-bit) weight CNNs are imported 
to an app on Android cellular phone (Samsung Galaxy J7) for 
runtime comparison. A user speaks one of the words on 
the list (`Yes', `No', `Up', `Down', `Left', `Right', `On', `Off', `Stop', `Go'). 
In Fig. \ref{nodemo}, a user speaks the work `No', the app recognizes it correctly and 
reports a runtime in millesecond (ms).  The full-precision model runs 120 ms while the binary weight 
model takes 59 ms, a 2x speed up on this particular hardware. A similar speed-up is observed on other words, for example the word `Off' shown in Fig. \ref{offdemo}.
While importing the binary weight CNN to the app, we keep the standard `conv2d' and `matmul' TensorFlow functions in the CNN. 
Additional speed-up is possible with a 
fast multiplication algorithm utilizing the binary weight structure 
and/or hardware consideration.

\begin{table}[htbp]
\caption{Test accuracies of the full precision (32 bit) and the binary weight keyword CNN by different training algorithms and 
initializations (cold/random start, warm start), without ($\rho=0$) and with blending ($\rho > 0$). }\label{ac1}
\centering
%%\begin{tabular}{p{100pt}p{100pt}p{100pt}p{100pt}p{100pt}p{100pt}}
%\begin{tabular}{p{65pt}p{65pt}p{65pt}p{65pt}p{65pt}p{%65pt}}
%\toprule
%start & 32 bit   & Median BC  & Median BR & BC  & BR % \\
%\midrule
%cold  &   87.9 \%    &   83.3 \%  &  83.4 \%  & 83.0 %\%  & 84.5 \%   \\
%\midrule
%warm  &  93.5 \%      &   92 \%    & 91.8 \%  &   %91.7 \%  & 92. 1 \% \\
%\midrule
%warm, blend &  & {\bf 92.4 \%}  &  92.1 \%   &  92.2 %\% & 91.6  \% \\ 
%\bottomrule
%\end{tabular}
%\end{table}
%
\begin{tabular}{p{65pt}p{65pt}p{65pt}p{65pt}p{65pt}}
\toprule
start & 32 bit   & Median BC  & BC  & BR  \\
\midrule
cold  &   87.9 \%    &   83.3 \%  & 83.0 \%  & 84.5 \%   \\
\midrule
warm  &  93.5 \%      &   92 \%   &   91.7 \%  & 92.1 \% \\
\midrule
warm, blend &  & {\bf 92.4 \%}   &  92.2 \% & 91.6  \% \\
\bottomrule
\end{tabular}
\end{table}
\medskip

We also tested median BC, BC and BR on CIFAR-10 color image dataset. In the experiments, we used the testing images for validation and warm start from pre-trained full precision models. We coded up the algorithms in PyTorch \cite{pytorch} platform and carried out the experiments on a desktop with NVIDIA GeForce GTX 1080 Ti. We ran 200 epochs for each test. The initial learning rate is 0.1 and decays by a factor of 0.1 at epochs 80 and 140. Phase II of BR starts at epoch 150 and the parameter $\lambda$ in BR increases by a factor of $1.02$ after every half epoch. In addition, we used batch size $= 128$, $\ell_2$ weight decay $=10^{-4}$ and momentum $= 0.95$. We tested the algorithms on VGG 
%\cite{vgg_14} and ResNet\cite{resnet_15} 
architectures \cite{vgg_14}, and the validation accuracies for CIFAR-10 are summarized in Table \ref{ac2}. We see that median BC outperforms BC, however blending appears to help BC and BR more than median BC in general. 

\begin{table}[htbp]
%\caption{Test accuracies of ResNet and VGG networks on CIFAR-10. }\label{ac2}
\caption{Test accuracies of VGG networks on CIFAR-10 image dataset. }\label{ac2}
\centering
%\begin{tabular}{p{90pt}p{50pt}p{65pt}p{65pt}p{65pt}p{55pt}}
\begin{tabular}{p{90pt}p{50pt}p{65pt}p{65pt}p{65pt}}
\toprule
%Network & 32 bit   & Median BC  & Median BR & BC  & BR 
Network & 32 bit   & Median BC  & BC  & BR
\\
\midrule
%ResNet20  &  92.42 \%      &  86.97  \%    & 89.24 \%  &   87.36 \%  & 89.69 \% \\
%\midrule
%ResNet20, blend &  &  87.27 \%  &  89.90 \%   &  88.72 \% & 90.77  \% \\
\midrule
%ResNet32  &  92.94 \%      &  89.30  \%    & 91.34 \%  &   89.17 \%  & 91.01 \% \\
%\midrule
%ResNet32, blend &  &  89.55 \%  &  91.67 \%   &  89.70 \% & 92.00 \% \\
%\midrule
%ResNet44  &  93.18 \%      &  89.92/90.28\%    &   &   90.48/90.39\%  & 91.51 \% \\
%\midrule
%ResNet44, blend &  &  90.22/90.04\%  &     &  90.25/91.10\% & 91.88 \% \\
%\midrule
%ResNet56  &  93.62 \%      &   90.44/90.72\%    &   &   90.82/91.21\%  & 91.82 \% \\
%\midrule
%ResNet56, blend &  &  90.72/90.50\%  &     &  91.43/90.97\% & 92.39 \% \\
%\midrule
%VGG-11  &  92.01 \%      &  89.49  \%    & 27.46 \%  &   88.03 \%  & 88.90 \% \\
VGG-11  &  92.01 \%      &  89.49  \%  &   88.03 \%  & 88.90 \% \\
\midrule
%VGG-11, blend &  &  90.02 \%  &  32.79 \%   &  89.35 \% & 89.51  \% \\
VGG-11, blend &  &  90.02 \%   &  89.35 \% & 89.51  \% \\
\midrule
VGG-13  &  93.56 \%      &   92.16 \%  &   91.85 \%  & 92.05 \% \\
\midrule
VGG-13, blend &  &  92.25 \%  &  91.85 \% &  92.48 \% \\
\midrule
%VGG-16  &  93.45 \%      &   91.83 \%    & 36.37 \%  &   91.68 \%  & 92.08 \% \\
VGG-16  &  93.45 \%      &   91.83 \%  &   91.68 \%  & 92.08 \% \\
\midrule
%VGG-16, blend &  &  92.01 \%  &  36.75 \%   &  92.21 \% &  92.74 \% \\

VGG-16, blend &  &  92.01 \%  &  92.21 \% &  92.74 \% \\

\bottomrule
\end{tabular}
\end{table}
\medskip

%\begin{table}[htbp]
%    \centering
%    \begin{tabular}{|c|c|c|}
%    \hline
%        $\rho$ & ResNet44 & ResNet56\\
%        \hline
%        1e-4 & 89.89 & 90.10 \\
%        5e-5 & 89.95 & 90.09 \\
%        4e-5 & 90.00 & 90.30 \\
%        3e-5 & 89.91 & 89.72 \\
%        2e-5 & 90.02 & 89.90 \\
%        1e-5 & 89.84 & 89.87 \\
%        5e-6 & 89.89 & 89.92 \\
%        4e-6 & 89.60 & 89.46 \\
%        3e-6 & 89.98 & 89.23 \\
%        2e-6 & 89.94 & 90.48 \\
%        1e-6 & 89.97 & 89.82 \\
%        0 & 90.48 & 90.24 \\\hline
%    \end{tabular}
%    \label{tab:my_label}
%\end{table}

\section{Conclusion}
We studied training algorithms of binary convolutional neural networks via the 
closed form $\ell_1$ projector on TensorFlow and PyTorch. The median operation takes place of 
arithmatic average in the standard $\ell_2$ projector. Under warm start, the median BinaryConnect improves the regular BinaryConnect on both audio and image CNNs. The trained binary CNN doubles the speed of 
full precision CNN when implemented on an Android phone app and tested on spoken keywords. 
\medskip

We observed that the blending technique in the $\ell_2$ projector context \cite{BCGD_18} tends to benefit BC more than median BC in experiments on both audio and image data. It remains to develop an alternative blending method for the median BC.  
\medskip

In future work, we plan to study $\ell_1$ projection in binary and higher-bit CNN training on 
larger datasets, and further understand its strengths.  The higher-bit exact and approximate projection formulas in the $\ell_1$ sense are derived in the appendix.

\medskip

\section*{Acknowledgements}
We would like to thank Dr. Meng Yu for suggesting the TensorFlow keyword CNN \cite{tf18} 
in this study, and helpful conversations while our work was in progress.

\medskip

%\bibliographystyle{spmpsci} 
%\bibliography{references}

\section*{Appendix: Median $m$-bit Projection Formulas ($m\geq 2$)}
\setcounter{equation}{0}
Let us consider $m=2$ and the problem of finding the closest ternary (2-bit) vector in the $\ell_1$ sense to a given 
real vector $w$, or the projection ${\rm proj}_{\Q} \, w$, 
for $w \in \R^D$, $\Q =\R_{+}\times \{0, \pm 1\}^D$.
\medskip

Write $z= s \, q$, where $s > 0$, $q=(q_j)$, $q_j= 0, \pm 1$. Let the index set of nonzero 
components of $z$ be $J$ with $t:=|J|$, the cardinality of $J$. Clearly, 
the $q_j=0$ ($j \not \in J$) approximates the $D-t$ smallest entries of $w$ in absolute value. 
So $J$ is the index set of $t$ largest components of $w$ in absolute value. 
Let $w_{[k]}$ extract the largest $k$ components of $w$ in absolute value, and zero out the rest. Then  
\be
\| z - w\|_{1} = \|w - w_{[t]}\|_1 + \sum_{j \in J}\; | s \, q_j - w_j |.  \label{ap1}
\ee
To minimize the expression in (\ref{ap1}), we must have $q_j = {\rm sgn}(w_j)$, $\forall j \in J$, sgn(0):=0, and:
\be
\| z - w\|_{1} = \|w - w_{[t]}\|_1 + \sum_{j \in J}\; | s \, - |w_j| \, |. \label{ap2}
\ee
The optimal value of $s$ is ${\rm median}(\{|w_j|: j \in J\})$. 
Let:
\be
t^* := {\rm argmin}_{1\leq t\leq D}\;  \|w - w_{[t]}\|_1 + \, 
\sum_{j \in J}\; |\, {\rm median}(\{|w_j|: j \in J\})  \, - |w_j| \,| 
\label{ap3}
\ee
and $J^*$ be the index set of $t^*$ largest components of $w$ in absolute value, 
then the optimal solution to (\ref{ap1}) is $z^* = s^*\, q^*$ with:
\be
s^*= {\rm median}(\{|w_j|: j \in J^* \}), \; 
q^{*}={\rm sgn}(w_{[t^*]}).
%
%q^{*}_{j} = {\rm sgn}(w_j), \; j \in J^*,\; \; q^{*}_{j}=0, \; \; j \not \in J^* 
\label{ap4}
\ee
%where ${\rm sgn}(0) = 0$.
\medskip

It is interesting to compare with the ternary projection in the $\ell_2$ sense \cite{tquant}:
\[ t^*= {\rm argmax}_{1\leq t \leq D}\, \| w_{[t]}\|_{1}^{2}/t, \; s^*= \|w_{[t^*]}\|_{1}/t^*,\; 
q^{*}={\rm sgn}(w_{[t^*]}). \] 
\medskip

For $m\geq 3$, exact solutions are combinatorial in nature and too costly computationally \cite{tquant}. A few iterations of 
Lloyd's algorithm \cite{Ld} is more effective \cite{BR_18,BCGD_18}, which iterates between the assignment step ($q$-update) 
and centroid step ($s$-update). In the $q$-update of the $l$-th iteration, with $s=s^{l-1}$ known from previous step, 
each component of $q^l$ is selected from the quantized state $\{\pm q_1,\cdots, \pm q_{m} \}$ so that 
$s^{l-1} q^l$ is nearest to $w$ component by component. In the $s$-update, the problem:
\be
 s^*={\rm argmin}_{s \in \R} \, \|\, s\,  q^l - w\, \|_{1} = {\rm argmin}_{s \in \R} 
\sum_{j: |q^{l}_{j}| > 0} \; |q^{l}_{j} |\; \|\, s - w_{j}/q^{l}_{j}\, \|_1
\label{ap5}
\ee
has closed form solution (eq. (2.3), \cite{sp16}). 
Let us rename the sequence $\{j: |q^{l}_{j} | > 0\}$ as $\{h_i\}_{i=1}^{I}$, where $I$ equals  
the cardinality of $\{j: |q^{l}_{j} | > 0\}$. 
Then:
\be
s^* = w_{j^*}/q^{l}_{j^*}  \label{ap6}
\ee
where $j^*$ corresponds to the renamed index 
\[ i^* = \min \left \{ k\in [1,I]: \sum_{i=1}^{I} h_i < 2 \sum_{i=1}^{k} h_{i} \right \}. \]

\end{document}